# Multisource Heterogeneous Domain Adaptation With Conditional Weighting Adversarial Network

Yuan Yao, Xutao Li, Yu Zhang, *Member, IEEE*, and Yunming Ye

*Abstract*— Heterogeneous domain adaptation (HDA) tackles the learning of cross-domain samples with both different probability distributions and feature representations. Most of the existing HDA studies focus on the *single-source* scenario. In reality, however, it is not uncommon to obtain samples from multiple heterogeneous domains. In this article, we study the *multisource* HDA problem and propose a conditional weighting adversarial network (CWAN) to address it. The proposed CWAN adversarially learns a feature transformer, a label classifier, and a domain discriminator. To quantify the importance of different source domains, CWAN introduces a sophisticated conditional weighting scheme to calculate the weights of the source domains according to the conditional distribution divergence between the source and target domains. Different from existing weighting schemes, the proposed conditional weighting scheme not only weights the source domains but also implicitly aligns the conditional distributions during the optimization process. Experimental results clearly demonstrate that the proposed CWAN performs much better than several state-of-the-art methods on four real-world datasets.

*Index Terms*— Adversarial network, conditional weighting, heterogeneous domain adaptation (HDA), multisource, semisupervised setting.

## Nomenclature

| | |
|---|---|
| $\mathcal{D}_{s_k}$ | $k$th source domain. |
| $\mathcal{D}_t$ | Target domain. |
| $\mathcal{D}_l/\mathcal{D}_u$ | Labeled/unlabeled target domain. |
| $\mathbf{x}_i^{s_k}$ | $i$th source sample in $\mathcal{D}_{s_k}$. |
| $\mathbf{x}_i^l/\mathbf{x}_i^u$ | $i$th labeled/unlabeled target sample. |
| $\mathbf{x}_{i,c}^{s_k}/\mathbf{x}_{i,c}^l$ | $i$th sample of class $c$ from $\mathcal{D}_{s_k}/\mathcal{D}_l$. |
| $\mathbf{y}_i^{s_k}/\mathbf{y}_i^l$ | Ground-truth one-hot class label of $\mathbf{x}_i^{s_k}/\mathbf{x}_i^l$. |
| $\widetilde{y}_{i,c}^u$ | Probability of $\mathbf{x}_i^u$ associated with class $c$. |
| $C$ | #classes (the number of the classes). |
| $d_{s_k}/d_t/d_c$ | Dimension of the features in $\mathcal{D}_{s_k}/\mathcal{D}_t$/common subspace. |
| $n_{s_k}/n_t/n_l/n_u$ | #samples in $\mathcal{D}_{s_k}/\mathcal{D}_t/\mathcal{D}_l/\mathcal{D}_u$. |
| $n_{s_k}^c/n_l^c$ | Total number of samples belonging to class $c$ in $\mathcal{D}_{s_k}/\mathcal{D}_l$. |
| $g(\cdot)/d(\cdot)/f(\cdot)$ | Feature transformer/domain discriminator/label classifier. |
| $g_{s_k}(\cdot)/g_t(\cdot)$ | Feature transformer in $\mathcal{D}_{s_k}/\mathcal{D}_t$. |
| $g_{s_k}^{l_1}(\cdot)/g_{s_k}^{l_2}(\cdot)$ | First/second layer of $g_{s_k}(\cdot)$. |
| $g_t^{l_1}(\cdot)/g_t^{l_2}(\cdot)$ | First/second layer of $g_t(\cdot)$. |
| $\delta_k$ | Dissimilarity between the $k$th source and target domains. |
| $w_k$ | Weight of the $k$th source domain. |
| $\mathcal{L}_g/\mathcal{L}_{d,g}/\mathcal{L}_{f,g}$ | Loss in $g(\cdot)/d(\cdot)/f(\cdot)$. |
| $\beta$ and $\tau$ | Parameters. |

Manuscript received August 14, 2020; revised February 9, 2021 and May 25, 2021; accepted August 10, 2021. This work was supported in part by the Shenzhen Science and Technology Program under Grant JCYJ20210324120208022, Grant JCYJ20180507183823045, and Grant JCYJ20200109113014456; and in part by NSFC under Grant 61972111, Grant U1836107, and Grant 62076118. *(Corresponding authors: Yunming Ye; Xutao Li.)*

Yuan Yao, Xutao Li, and Yunming Ye are with the Department of Computer Science and Technology, Harbin Institute of Technology, Shenzhen 518055, China (e-mail: yaoyuan@stu.hit.edu.cn; lixutao@hit.edu.cn; yeyunming@hit.edu.cn).

Yu Zhang is with the Department of Computer Science and Engineering, Southern University of Science and Technology, Shenzhen 518055, China, and also with the Peng Cheng Laboratory, Shenzhen 518066, China (e-mail: yu.zhang.ust@gmail.com).

Color versions of one or more figures in this article are available at https://doi.org/10.1109/TNNLS.2021.3105868.

Digital Object Identifier 10.1109/TNNLS.2021.3105868

## I. Introduction

HETEROGENEOUS domain adaptation (HDA) [1] aims to assist the learning task in an interesting but label-scarce domain, i.e., target domain, by leveraging the knowledge from a heterogeneous but label-rich domain, i.e., source domain. HDA techniques have been successfully applied to various real-world applications, such as cross-modality image classification [2]–[7] and cross-lingual text categorization [8]–[12]. However, all the methods focus on the *single-source* scenario, in which the source samples are collected from a single heterogeneous domain. Thus, they cannot be directly used for the *multisource* scenario, where the source samples are obtained from multiple related but heterogeneous domains. Note that the multiple source domains are heterogeneous not only from the target one but also from each other. This scenario is common and important in many real-world applications. For instance, in some natural language processing applications, the articles may be written by different languages (e.g., English, Japanese, and Chinese); hence, their feature representations are *heterogeneous* since different vocabularies are utilized (e.g., the top row in Fig. 1) [13]. Similarly, in some industrial applications, the important samples may be protected by a privacy policy (e.g., the EU General Data Protection Regulation (GDPR) [14]). Also, users do not want their private information (e.g., photographs) to be leaked [15]. Although the modern multinode learning system (e.g., distributed learning [16], [17] and federated learning [18]) has recently emerged to tackle this challenge, the private training samples still have the risk of leakage through gradients exchange [19]. To avoid such leakage, a possible solution is to utilize various preprocessed features (e.g., SURF [20], DeCAF$_6$ [21], and ResNet$_{50}$ [22]) provided by different participants rather than the raw samples (e.g., photographs) for training. Those features are quite hard to recover raw samples because the parameters and architectures of feature extractors are not provided for privacy protection. Also, those features are often *heterogeneous* since different participants usually use different feature extractors for data security (e.g., the bottom row in Fig. 1). Hence, it is highly desirable and meaningful to transfer knowledge across multiple heterogeneous domains, especially for the privacy-protected samples. Also, as multiple heterogeneous domains may offer complementary knowledge, a well-designed model will be expected to effectively utilize that knowledge for better transfer performance. This thinking is in line with the spirit of the multisource HDA (MHDA) problem. Unfortunately, the MHDA problem is not well studied.

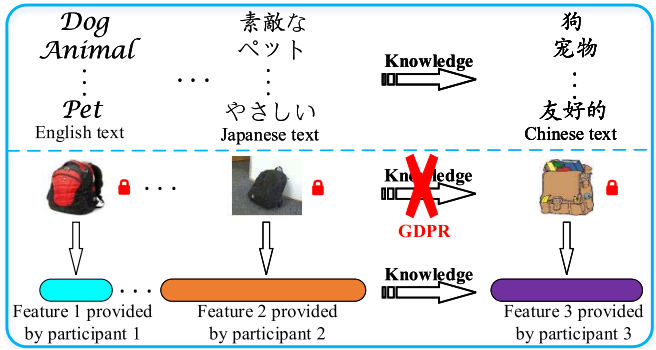

Fig. 1. Example scenarios for MHDA. Top row: the target articles are written in Chinese, while the source articles are written in multiple different languages, such as English and Japanese. Bottom row: the source and target samples are characterized by various features for privacy protection, which are quite hard to recover raw samples because the parameters and architectures of feature extractors are not provided. Also, those features are often *heterogeneous* since different participants usually use different feature extractors for data security.

In the literature, there are very few approaches. One representative is domain adaptation using manifold alignment (DAMA) [23]. The method learns multiple linear transformations by simultaneously aligning the manifold structure of each domain and maximizing their discriminative contributions. However, it has two important drawbacks. First, DAMA treats each source domain equally and does not distinguish its importance. Second, shallow linear transformations are limited to tackle complex problems. Though recently several deep learning techniques have been developed for the multisource domain adaptation [24]–[27], they make a strong assumption that all the domains have the same feature representation. Hence, they are inapplicable to the MHDA problem.

In this article, we propose a conditional weighting adversarial network (CWAN) to solve the MHDA problem. In the CWAN architecture, a two-layer nonlinear feature transformer is constructed for each individual source domain and the target domain, which projects the source and target samples into an intermediate subspace. Upon the transformation, a domain discriminator is appended. By playing a minimax game, the feature transformer and the domain discriminator are simultaneously trained, which aligns the distributions of source domains with that of the target domain. To maintain the discriminative ability of the feature transformer, a label classifier is also incorporated and trained. Moreover, to balance the contribution of each source domain, we carefully model the conditional distribution divergence between the source and target domains in the intermediate subspace and design a sophisticated conditional weighting scheme.

The contributions of this article are summarized as follows.
1) We propose an adversarial network called CWAN, which, to the knowledge of our best, is the first deep learning model for the MHDA problem.
2) In CWAN, a sophisticated conditional weighting scheme is developed, which not only skillfully quantifies the importance of different source domains but also aligns the conditional distributions across the source and target domains for knowledge transfer. To the best of our knowledge, there is no existing weighting scheme to achieve both effects simultaneously.
3) Extensive experimental results are reported, which verify the proposed CWAN outperforms existing competitors.

The remainder of this article is organized as follows. In Section II, we first review related work. Then, we present the proposed CWAN in Section III. Next, in Section IV, we report extensive experimental results. Finally, we make conclusions and give several suggestions for future work in Section V.

## II. RELATED WORK

In this section, we first review several lines of study that are closely relevant to ours: single-source HDA (SHDA), MHDA, single-source *homogeneous* domain adaptation (SDA), and multisource *homogeneous* domain adaptation (MDA). Then, we highlight the difference between those related studies and ours.

### A. Single-Source HDA

Existing approaches generally bridge two heterogeneous domains by either deriving a domain-invariant subspace [3], [4], [6], [8], [10], [28]–[34] or transforming samples from one domain to another [5], [9], [11], [12], [35], [36]. Those approaches are mainly designed according to the classifier adaptation and distribution alignment strategies. For example, Li *et al.* [8], Duan *et al.* [28], and Hoffman *et al.* [35], [36] leverage the former to align the discriminative structures of the source and target domains, Yan *et al.* [11] and Li *et al.* [30] adopt the latter to match the distributions of the two domains, and Tsai *et al.* [5], Xiao and Guo [9], Hsieh *et al.* [10], Yan *et al.* [29], and Yao *et al.* [33] employ both for combining their strengths. However, these approaches are shallow learning models, which cannot tackle complex scenarios. Recently, several studies [3], [4], [6], [31] have turned to deep learning techniques for the SHDA problem. Specifically, Shu *et al.* [3] propose a weakly shared strategy to minimize the difference of the parameters in the last layers of the source and target projection networks. Chen *et al.* [4] propose a transfer neural tree (TNT). This method simultaneously deals with feature projection, adaptation, and categorization in a unified framework. Li *et al.* [31] put forward a deep matrix completion with adversarial kernel embedding (Deep-MCA). This method builds a deep neural network to complete the heterogeneous feature matrix and employs the idea of adversarial learning to seek a better measure function for distribution alignment. Yao *et al.* [6] present a soft transfer network (STN). This method constructs, respectively, a two-layer transformation to project the samples of source and target domains into a common subspace. In order to minimize the conditional distribution divergence, a soft-label strategy and an iterative weighting scheme are developed. Upon the projection, STN appends a label classifier. More recently, Wang *et al.* [37] propose a graph learning method to not only minimize the gap across domains but also maximize the margin among distinct classes. Yao *et al.* [33] develop a discriminative distribution alignment (DDA) framework. This framework simultaneously performs classifier adaptation, distribution alignment, and discriminative embedding. Li *et al.* [38] present a simultaneous semantic alignment network (SSAN). This method takes both implicit and explicit semantic alignments into account. However, in the MHDA problem, as the given multiple source domains cannot be merged into a larger source domain because of the heterogeneous feature representations, the SHDA methods are inapplicable.

### B. Multisource HDA

There are very few approaches to study the MHDA problem. The DAMA [23] is a representative. As noted in Section I, however, DAMA neither distinguishes the importance of different source domains nor utilizes deep learning techniques.

### C. Single-Source Homogeneous Domain Adaptation

As many methods have been presented to tackle the type of problem [39]–[42], we just review the studies closely related to our work. In [43]–[47], the *pseudo-label* strategy is applied to align the conditional distributions by minimizing the class-conditional maximum mean discrepancy (MMD) [48]. Another line of studies utilizes deep adversarial learning techniques [49]–[58] to match distributions of the source and target domains. However, in those methods, feature transformers of the source and target domains either share the same network or adopt two identical networks with a parameter consistency constraint. This fact implies that they cannot deal with heterogeneous inputs. Hence, these approaches are not readily applicable to SHDA, not to mention MHDA.

### D. Multisource Homogeneous Domain Adaptation

Yang *et al.* [59] first propose an adaptive support vector machine (A-SVM), which trains a classifier in each domain and then employs a weighted ensemble scheme to combine them. To reduce the gap between the weighted prediction and target ground domain truth, an SVM is learned. Inspired by the study, many shallow learning approaches [60]–[66] are developed for the MDA problem. Recently, researchers have resorted to deep learning techniques for this problem [24]–[27], [67]. However, these methods utilize a shared feature transformer for multiple source domains and the target domain. Thus, they cannot tackle the MHDA problem.

### E. Discussions

In this work, we focus on addressing the MHDA problem, which substantially distinguishes from other related studies on the following grounds.

1) SHDA methods can only handle two heterogeneous inputs. When more heterogeneous source domains are involved, they need to introduce new learnable parameters (i.e., feature transformations), which leads to complex optimization problems. In addition, they are obviously incapable of distinguishing the importance of different source domains.
2) SDA methods can only tackle homogeneous inputs, which can easily be extended to the MDA problem. However, similar to SHDA methods, when solving the MHDA problem, they also need to involve new learnable parameters and cannot quantify the contribution of each source domain. In addition, most SDA methods focus on tackling unsupervised settings where no target labels are provided, while MHDA methods focus on handling semisupervised settings in which a few target labels are available.
3) MDA methods can only deal with homogeneous inputs. Most of them can calculate the contributions of distinct source domains to the target one. However, similar to SDA and SHDA methods, they have to solve new learnable parameters when involving multiple heterogeneous source domains. Moreover, MDA and MHDA methods focus on dealing with unsupervised and semisupervised settings, respectively.

## III. CONDITIONAL WEIGHTING ADVERSARIAL NETWORK

In this section, we begin by introducing the problem formulation and notations. Then, we present the proposed CWAN. Finally, we discuss the difference between the proposed weighting scheme and several related ones.

### A. Problem Formulation and Notations

In the MHDA problem, we are given $K$ heterogeneous source domains and a target domain. Let $\mathcal{D}_{s_k} = \{(\mathbf{x}_i^{s_k}, \mathbf{y}_i^{s_k})\}_{i=1}^{n_{s_k}}$ be the $k$th source domain, where the $i$th sample $\mathbf{x}_i^{s_k}$ is represented by $d_{s_k}$-dimensional features, and $\mathbf{y}_i^{s_k}$ is the corresponding one-hot class label over $C$ classes. Analogously, the target domain is denoted by $\mathcal{D}_t = \mathcal{D}_l \cup \mathcal{D}_u = \{(\mathbf{x}_i^l, \mathbf{y}_i^l)\}_{i=1}^{n_l} \cup \{\mathbf{x}_i^u\}_{i=1}^{n_u}$, where $\mathbf{x}_i^l$ ($\mathbf{x}_i^u$) is the $i$th labeled (unlabeled) target domain sample with $d_t$-dimensional features, and $\mathbf{y}_i^l$ is its associated one-hot class label over $C$ classes. For convenience, we let $\{\mathbf{x}_i^t\}_{i=1}^{n_t} = \{\{\mathbf{x}_i^l\}_{i=1}^{n_l}, \{\mathbf{x}_i^u\}_{i=1}^{n_u}\}$ denote all the samples in the target domain. As our problem is under the MHDA setting, we have $d_{s_1} \neq \cdots \neq d_{s_K} \neq d_t$, $\{n_{s_k} \gg n_l\}_{k=1}^{K}$, and $n_u \gg n_l$. The goal is to design a heterogeneous adaptation network for categorizing the samples in $\mathcal{D}_u$. For easy reference, we summarize the notations used in this article in Nomenclature.

### B. CWAN

The MHDA problem has three key challenges: 1) the features are heterogeneous in all the domains; 2) the distributions between each source and target domains are different; and 3) each source domain has a distinct contribution to the target domain, which is unknown. These challenges motivate the development of the CWAN, as depicted in Fig. 2. Here, a feature transformer $g(\cdot)$ is used to eliminate the feature heterogeneity. Upon the feature transformer, we introduce a domain discriminator $d(\cdot)$ and a label classifier $f(\cdot)$, which are adversarially trained for reducing the distributional divergence. In addition, the contribution of each source domain is calculated by developing a sophisticated conditional weighting scheme, marked with pink color. Next, we elaborate on how each part works and formulate the overall objective of CWAN.

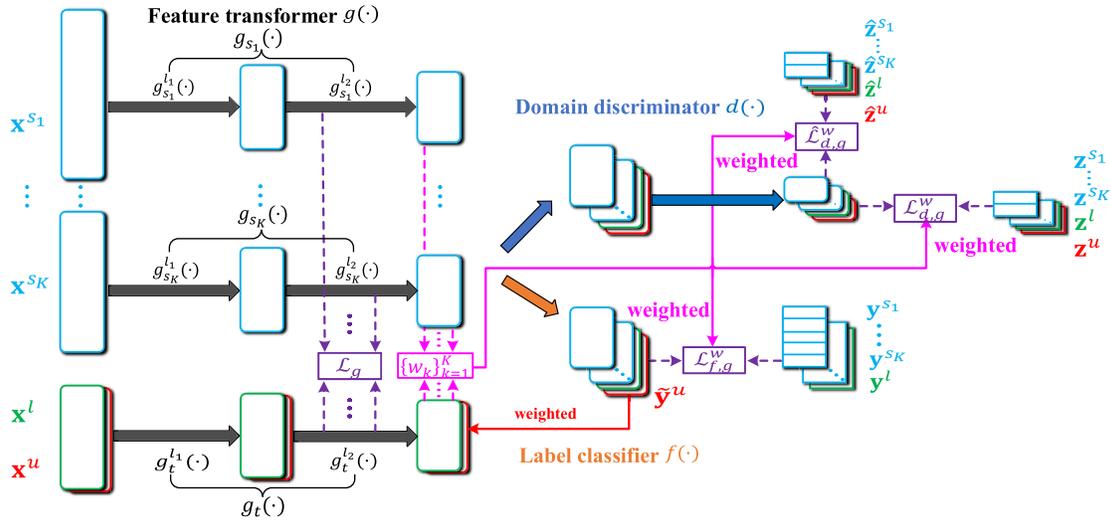

Fig. 2. Architecture of CWAN, which consists of a feature transformer $g(\cdot)$ that includes multiple source mapping networks $\{g_{s_k}(\cdot)\}_{k=1}^K$, a target mapping network $g_t(\cdot)$, a label classifier $f(\cdot)$, and a domain discriminator $d(\cdot)$. Here, $\{(\mathbf{x}^{s_k}, \mathbf{y}^{s_k})\}_{k=1}^K$ (in blue) and $\{(\mathbf{x}^l, \mathbf{y}^l)\}$ (in green) are labeled source and target samples, respectively, $\mathbf{x}_u$ (in red) is unlabeled target samples, $\widetilde{\mathbf{y}}_u = f(g_t(\mathbf{x}^u))$ (in red) is the soft-label of $\mathbf{x}^u$, $\{(\{\mathbf{z}^{s_k}\}_{k=1}^K, \mathbf{z}^l, \mathbf{z}^u)\}$ and $\{(\{\widehat{\mathbf{z}}^{s_k}\}_{k=1}^K, \widehat{\mathbf{z}}^l, \widehat{\mathbf{z}}^u)\}$ are true and inverted domain labels of $\{(\{\mathbf{x}^{s_k}\}_{k=1}^K, \mathbf{x}^l, \mathbf{x}^u)\}$, respectively, and $\{w_k\}_{k=1}^K$ (in pink) are the weights of the source domains. Our model contains four losses: $\mathcal{L}_g$ to model the correlations between domains, $\mathcal{L}_{f,g}^w$ for maintaining the discriminative ability of the feature transformer, and $\mathcal{L}_{d,g}^w$ and $\widehat{\mathcal{L}}_{d,g}^w$ to align the distributions of the source and target domains by adversarially training.

*1) Heterogeneous Feature Transformation:* As shown in Fig. 2, the feature transformer $g(\cdot)$ comprises multiple two-layer neural networks $\{g_{s_k}(\cdot):\mathbb{R}^{d_{s_k}} \to \mathbb{R}^{d_c}\}_{k=1}^K$ and $g_t(\cdot):\mathbb{R}^{d_t} \to \mathbb{R}^{d_c}$, which projects the samples of source and target domains into a $d_c$-dimensional subspace. Since the networks are built independently, the projections take only the domain-specific characteristics into account but fail to model the correlations between the source and target domains. One nature remedy is imposing a parameter consistency constraint on the projection networks. However, because of the diverse structures, we are unable to incorporate it directly. Thus, we assume that the second layers of transformations $\{g_{s_k}^{l_2}(\cdot)\}_{k=1}^K$ and $g_t^{l_2}(\cdot)$ have identical structures. Note that the assumption cannot be made for first layers of transformations $\{g_{s_k}^{l_1}(\cdot)\}_{k=1}^K$ and $g_t^{l_1}(\cdot)$ due to the heterogeneity of input features. Moreover, we expect that the diverse structures in the first layers can preserve the domain-specific characteristics. Then, we employ an $\ell_1$-norm to measure the disagreement between parameters in the second layers of projections as

$$\mathcal{L}_g = \sum_{k=1}^K \|g_{s_k}^{l_2} - g_t^{l_2}\|_1. \quad (1)$$

Minimizing the loss can control the disagreement between the source and target parameters, and it makes the parameters as consistent as possible. Accordingly, this scheme can flexibly model the correlations between the source and target domains, which is extremely practical in some complex situations.

In addition, heterogeneous feature transformation is the main difference between MDA and MHDA approaches. Compared to homogeneous feature transformation used in MDA approaches, heterogeneous feature transformation has the following advantages.

1) Heterogeneous feature transformation can handle multimodality samples simultaneously.
2) If the input modality changes, the architecture of heterogeneous feature transformation does not need to change.
3) Heterogeneous feature transformation can deal with different deep features (e.g., DeCAF$_6$ and ResNet$_{50}$) at the same time. Since those deep features preserve the most discriminative information of the samples, heterogeneous feature transformation does not need to stack many feature extraction layers that can achieve good performance, which is flexible and efficient.
4) Heterogeneous feature transformation is more general than homogeneous feature transformation.

*2) Adversarial Distribution Alignment:* Recently, domain adversarial networks [49] have been successfully applied in distribution alignment, and they build two competitive systems, i.e., the domain discriminator and the feature transformer. The former aims to distinguish the source and target samples, and the latter tries to fool the former. In such a competitive learning manner, the distributions of source and target domains can be aligned. Here, we follow the similar idea to design our architecture. The domain discriminator $d(\cdot)$ and the feature transformer $g(\cdot)$ form a competitive loss $\mathcal{L}_{d,g}$ in (2), which is to minimize over $d(\cdot)$ but maximize over $g(\cdot)$

$$\mathcal{L}_{d,g} = \sum_{k=1}^K \frac{1}{n_{s_k}} \sum_{i=1}^{n_{s_k}} L_s\big[d\big(g_{s_k}(\mathbf{x}_i^{s_k})\big), \mathbf{z}_i^{s_k}\big]$$
$$+ \frac{1}{n_t} \sum_{i=1}^{n_t} L_s\big[d\big(g_t(\mathbf{x}_i^t)\big), \mathbf{z}_i^t\big] \quad (2)$$

where $L_s[\cdot, \cdot]$ is the squared loss, and $\mathbf{z}_i^{s_k}$ and $\mathbf{z}_i^t$ are the one-hot domain labels of $\mathbf{x}_i^{s_k}$ and $\mathbf{x}_i^t$, respectively. In addition, a classification loss on the label classifier $f(\cdot)$ and feature transformer $g(\cdot)$ is also designed as

$$\mathcal{L}_{f,g} = \sum_{k=1}^{K} \frac{1}{n_{s_k}} \sum_{i=1}^{n_{s_k}} L_c\big[f\big(g_{s_k}(\mathbf{x}_i^{s_k})\big), \mathbf{y}_i^{s_k}\big]$$
$$+ \frac{1}{n_l} \sum_{i=1}^{n_l} L_c\big[f\big(g_t(\mathbf{x}_i^l)\big), \mathbf{y}_i^l\big] + \tau\big(\|f\|^2 + \|g\|^2\big) \quad (3)$$

where $L_c[\cdot, \cdot]$ is the cross-entropy loss and $\tau$ is a positive regularization parameter. The loss is minimized over both $f(\cdot)$ and $g(\cdot)$. The label classifier can be trained by optimizing $f(\cdot)$. On the other hand, minimizing $\mathcal{L}_{f,g}$ over $g(\cdot)$ leads to a better discriminability for the feature transformer. Putting the two losses together, we have

$$\min_{f,g} \max_{d} \mathcal{L}_{f,g} - \beta \mathcal{L}_{d,g} \quad (4)$$

where $\beta$ is the tradeoff parameter between the label classifier and the domain discriminator. Though the above objective can align the distributions of the source and target domains, due to the neglect of label information in $\mathcal{L}_{d,g}$, only the marginal distributions are matched. We will address the deficiency when designing the conditional weighting scheme next.

*3) Multisource Conditional Weighting:* In this section, we introduce how to design the conditional weighting scheme, which can not only quantify the contribution of each source domain but also implicitly align the conditional distributions. Specifically, we first reformulate $\mathcal{L}_{d,g}$ and $\mathcal{L}_{f,g}$ as

$$\mathcal{L}_{d,g}^w = \sum_{k=1}^{K} \frac{w_k}{n_{s_k}} \sum_{i=1}^{n_{s_k}} L_s\big[d\big(g_{s_k}(\mathbf{x}_i^{s_k})\big), \mathbf{z}_i^{s_k}\big]$$
$$+ \frac{1}{n_t} \sum_{i=1}^{n_t} L_s\big[d\big(g_t(\mathbf{x}_i^t)\big), \mathbf{z}_i^t\big] \quad (5)$$

$$\mathcal{L}_{f,g}^w = \sum_{k=1}^{K} \frac{w_k}{n_{s_k}} \sum_{i=1}^{n_{s_k}} L_c\big[f\big(g_{s_k}(\mathbf{x}_i^{s_k})\big), \mathbf{y}_i^{s_k}\big]$$
$$+ \frac{1}{n_l} \sum_{i=1}^{n_l} L_c\big[f\big(g_t(\mathbf{x}_i^l)\big), \mathbf{y}_i^l\big] + \tau\big(\|f\|^2 + \|g\|^2\big) \quad (6)$$

where $w_k$ is the weight of the $k$th source domain. It is easy to see that $\mathcal{L}_{d,g}$ and $\mathcal{L}_{f,g}$ are special cases of $\mathcal{L}_{d,g}^w$ and $\mathcal{L}_{f,g}^w$, respectively, when all $w_k$'s become one. Then, in principle, the more dissimilar the $k$th source and target domains are, the smaller $w_k$ is. As we also expect to match the conditional distributions, the dissimilarity is characterized by the divergence between conditional distributions. The class-conditional MMD [6] with linear kernel has been proven to be an effective tool for measuring the conditional distribution divergence across heterogeneous domains. Also, it is a nonparametric distance estimate between conditional distributions. Hence, the dissimilarity between the $k$th source and target domains is calculated as

$$\delta_k = \frac{1}{C} \sum_{c=1}^{C} \left\| \frac{\sum_{i=1}^{n_l^c} g_t(\mathbf{x}_{i,c}^l) + \sum_{i=1}^{n_u} \widetilde{y}_{i,c}^u g_t(\mathbf{x}_i^u)}{n_l^c + \sum_{i=1}^{n_u} \widetilde{y}_{i,c}^u} \right.$$
$$\left. - \frac{1}{n_{s_k}^c} \sum_{i=1}^{n_{s_k}^c} g_{s_k}(\mathbf{x}_{i,c}^{s_k}) \right\|^2 \quad (7)$$

where $\mathbf{x}_{i,c}^{s_k}$ and $\mathbf{x}_{i,c}^l$ are the $i$th samples of class $c$ from the $k$th source and target domains, respectively, $n_{s_k}^c$ and $n_l^c$ are the total number of samples belonging to class $c$ in the $k$th source and target domains, respectively, and $\widetilde{y}_{i,c}^u$ is the probability of $\mathbf{x}_i^u$ associated with class $c$ provided by $f(\cdot)$. Next, in order to meet our weighting principle, one naive approach is to utilize a monotone decreasing function, e.g., $h(\cdot) = 1/(1 + \exp(\cdot))$, to calculate $w_k$, i.e., $w_k = h(\delta_k)$. In this manner, however, as $w_k$ and $\delta_k$ are composite functions on $f(\cdot)$ and $g(\cdot)$, when optimizing them, $w_k$ will be minimized, leading to the maximization of $\delta_k$. Thus, to avoid this, we utilize a monotonically increasing function to calculate $w_k$

$$w_k = \frac{1}{K-1} \left( \sum_{j=1}^{K} \frac{\exp(\delta_j)}{1 + \exp(\delta_j)} - \frac{\exp(\delta_k)}{1 + \exp(\delta_k)} \right)$$
$$= \frac{1}{K-1} \sum_{\substack{j=1 \\ j \neq k}}^{K} \frac{\exp(\delta_j)}{1 + \exp(\delta_j)} \quad (8)$$

where $(\exp(\delta_j))/(1 + \exp(\delta_j))$ is used to scale $\delta_j$ from $[0, +\infty)$ to $[0.5, 1)$, which leads to $w_k \in [0.5, 1)$ for better weighting. Note that the value of $w_k$ depends on those of $\{\delta_j\}_{j=1, j \neq k}^K$, not that of $\delta_k$. Moreover, the minimum value of $w_k$ is 0.5 instead of 0. Thus, this scheme does not discard any source domains but only reduces the importance of the dissimilar source domains. On the one hand, if the $k$th source domain is more dissimilar to the target domain than other source domains, then $\{\delta_k > \delta_j\}_{j=1, j \neq k}^K$, leading to $\{w_k < w_j\}_{j=1, j \neq k}^K$, and vice versa. On the other hand, $\{w_k\}_{k=1}^K$ and $\{\delta_k\}_{k=1}^K$ are a set of composite functions on $f(\cdot)$ and $g(\cdot)$. Thus, when optimizing them, $\{w_k\}_{k=1}^K$ will be minimized, which leads to the minimization of $\{\delta_j\}_{j=1}^K$. As a result, the conditional distributions are aligned.

*4) Overall Objective of CWAN:* Combing the three parts, we have the overall optimization function of CWAN

$$\min_{g,f} \max_{d} \mathcal{L}_{f,g}^w + \mathcal{L}_g - \beta \mathcal{L}_{d,g}^w. \quad (9)$$

However, the adversarial optimizations on the $-\mathcal{L}_{d,g}^w$ term have two important defects: 1) it may prevent the minimization of $\{w_k\}_{k=1}^K$ and 2) it may lead to vanishing gradients if $g(\cdot)$ and $d(\cdot)$ are not carefully synchronized. To avoid these problems, we adopt the invert label loss [50], [68] and define

$$\widehat{\mathcal{L}}_{d,g}^w = \sum_{k=1}^{K} \frac{w_k}{n_{s_k}} \sum_{i=1}^{n_{s_k}} L_s\big[d\big(g_{s_k}(\mathbf{x}_i^{s_k})\big), \widehat{\mathbf{z}}_i^{s_k}\big]$$
$$+ \frac{1}{n_t} \sum_{i=1}^{n_t} L_s\big[d\big(g_t(\mathbf{x}_i^t)\big), \widehat{\mathbf{z}}_i^t\big] \quad (10)$$

where $\widehat{\mathbf{z}}_i^{s_k}$ and $\widehat{\mathbf{z}}_i^t$ are the one-hot inverted domain labels of $\mathbf{x}_i^{s_k}$ and $\mathbf{x}_i^t$, respectively. Accordingly, the objective in (9) can be reformulated as

$$\min_{g,f} \mathcal{L}_{f,g}^w + \mathcal{L}_g + \beta \widehat{\mathcal{L}}_{d,g}^w, \quad \min_{d} \mathcal{L}_{d,g}^w. \quad (11)$$

By optimizing (11), we can learn the optimized feature transformer $g(\cdot)$, the domain discriminator $d(\cdot)$, and the label classifier $f(\cdot)$.

TABLE I
STATISTICS OF THE DATASETS. HERE, $|N|$, $|F|$, AND $|C|$ DENOTE THE NUMBER OF SAMPLES, FEATURES, AND CLASSES, RESPECTIVELY; $|\text{SURF}|$, $|\text{DECAF}_6|$, AND $|\text{RESNET}_{50}|$ DENOTE THE FEATURE DIMENSIONS OF SURF ($S_{800}$), DECAF$_6$ ($D_{4096}$), AND RESNET$_{50}$ ($R_{2048}$), RESPECTIVELY

| Dataset | Type | Domain | $|N|$ | $|F|$ | $|C|$ |
|---|---|---|---|---|---|
| Multilingual Returns Collection | Text | English (E) | 18758 | 1131 | 6 |
| | | French (F) | 26648 | 1230 | |
| | | German (G) | 29953 | 1417 | |
| | | Italian (I) | 24039 | 1041 | |
| | | Spanish (S) | 12342 | 807 | |
| Office-Home | Image | Artistic (Ar) | 2427 | $|\text{SURF}|$: 800, $|\text{DECAF}_6|$: 4096, $|\text{ResNet}_{50}|$: 2048 | 65 |
| | | Clip-art (Cl) | 4365 | | |
| | | Product (Pr) | 4439 | | |
| | | Real-world (Re) | 4357 | | |
| Office-31 | Image | Amazon (A) | 2817 | $|\text{SURF}|$: 800, $|\text{DECAF}_6|$: 4096, $|\text{ResNet}_{50}|$: 2048 | 31 |
| | | Webcam (W) | 795 | | |
| | | DSLR (D) | 498 | | |
| ImageNet+NUS-WIDE | Image, Text | Image (Im) | 800 | 4096 | 8 |
| | | Image-noise (In) | 800 | 600 | |
| | | Text (Te) | 800 | 64 | |

## C. Comparison With Existing Weighting Schemes

We now compare the proposed conditional weighting scheme with some existing studies. To the best of our knowledge, the most closely related multisource weighting schemes are presented in [60] and [61] for the MDA problem. However, the proposed multisource conditional weighting scheme substantially distinguishes from them in the following aspects.
1) Duan et al. [60] assign the weight to each source domain by measuring the divergence between marginal distributions rather than between conditional distributions.
2) Although Chattopadhyay et al. [61] consider the conditional distribution divergence, it estimates the weights based on a smooth manifold assumption. However, the assumption hinders its applications in reality.
3) The weighting schemes in [60] and [61] lack the ability to align the conditional distributions across domains, which is very important for effective knowledge transfer.

## D. Discussion on the Transferability of the Feature-Level

In the MHDA problem, we often encounter the situation where only the extracted features are available, especially for the privacy-protected samples. Thus, it is necessary to explore the transferability of the feature level. The extracted features are transferable mainly due to the following two guarantees.
1) The extracted features can preserve some information of the raw samples. If the source and target samples are related, then the features extracted from them are also related to a certain extent, which is a guarantee.
2) The class labels of source features and a small number of target features are available, which can be used to train a domain-shared classifier and align the conditional distributions between the source and target domains. According to the studies in [5], [6], [10], and [33], we find that both strategies are useful and important for HDA. Thus, the available label information is another guarantee.

## IV. EXPERIMENTS

In this section, we perform extensive experiments to verify the effectiveness of the proposed CWAN. The codes and datasets are available at https://github.com/yyyaoyuan/CWAN.

## A. Setup

*1) Datasets:* We empirically evaluate the proposed CWAN on four real-world datasets: **Multilingual Reuters Collection** [69], **Office-Home** [70], **Office-31** [71], and **ImageNet + NUS-WIDE** [72], [73]. The statistics of the datasets are listed in Table I.

The **Multilingual Reuters Collection** dataset comprises over 11 000 articles from six classes in five languages, i.e., English (**E**), French (**F**), German (**G**), Italian (**I**), and Spanish (**S**). Following Tsai et al. [5], Yao et al. [6], Li et al. [8], and Hsieh et al. [10], we represent each article by bag-of-words (BOW) with TF-IDF features and then perform dimension reduction using PCA with 60% energy preserved. The final dimensions w.r.t. **E**, **F**, **G**, **I**, and **S** are 1131, 1230, 1417, 1041, and 807, respectively. We treat **S** as the target domain and any two of the remaining languages as the source domains. Thus, we obtain six transfer tasks. For each source domain, we randomly choose 100 articles per class as the labeled samples. As for the target domain, we randomly pick up five and 500 articles from each category as the labeled and unlabeled samples, respectively.

The **Office-Home** dataset includes 15 500 images of 65 categories collected from four dissimilar domains: Artistic (**Ar**), Clip-art (**Cl**), Product (**Pr**), and Real-world (**Re**). Some sample images of the category of *Alarm_Clock* are illustrated in Fig. 3. We consider **Re** as the target domain and any two of the other domains as the source ones. We describe each image with three kinds of features: 800-D SURF ($S_{800}$), 4096-D DeCAF$_6$ ($D_{4096}$), and 2048-D ResNet$_{50}$ ($R_{2048}$). We first design three groups of transfer directions: $D_{4096}$, $R_{2048} \to S_{800}$; $S_{800}$, $R_{2048} \to D_{4096}$; and $S_{800}$, $D_{4096} \to R_{2048}$. Then, for each group, we construct three transfer tasks: **Ar**, **Cl** $\to$ **Re**; **Ar**, **Pr** $\to$ **Re**; and **Cl**, **Pr** $\to$ **Re**, leading to nine transfer tasks in total. In addition, we use all images in each source domain as the labeled source samples and randomly pick up three images per category from the target domain as the labeled target samples. The remaining images in the target domain are used as the unlabeled target samples.

The **Office-31** dataset consists of over 4000 images of 31 objects in three different domains: Amazon (**A**), Webcam (**W**), and DSLR (**D**). Some sample images of the category of *Back_Pack* are depicted in Fig. 4. We represent each image using the above mentioned features, i.e., $S_{800}$, $D_{4096}$, and $R_{2048}$. Similar to the *Office-Home* dataset, we first assign three sets of transfer directions: $D_{4096}$, $R_{2048} \rightarrow S_{800}$; $S_{800}$, $R_{2048} \rightarrow D_{4096}$; and $S_{800}$, $D_{4096} \rightarrow R_{2048}$. Then, we build three transfer tasks for each set: A, D $\rightarrow$ W; A, W $\rightarrow$ D; and W, D $\rightarrow$ A, which leads to nine transfer tasks in total. For each source domain, we take all the images as the labeled samples. As for the target domain, we randomly select three images per class as the labeled samples, and the other images are used as the unlabeled samples.

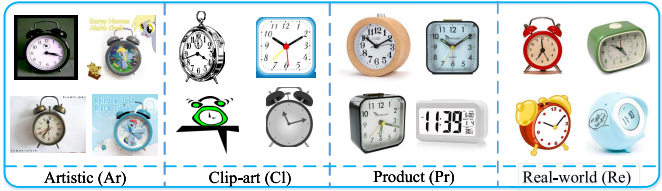

Fig. 3. Some sample images of the *Alarm_Clock* category from the Office-Home dataset. It contains four dissimilar domains: Artistic (Ar), Clip-art (Cl), Product (Pr), and Real-world (Re).

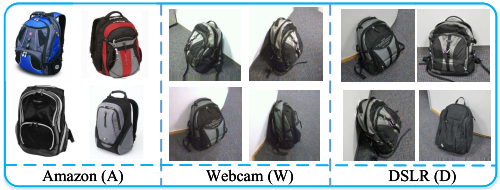

Fig. 4. Some sample images of the *Back_Pack* category from the Office-31 dataset. It includes three distinct domains: Amazon (A), Webcam (W), and DSLR (D).

The **ImageNet + NUS-WIDE** dataset contains two sub-datasets, i.e., ImageNet [72] and NUS-WIDE [73]. Following Chen *et al.* [4], we choose eight common classes among them to construct the ImageNet + NUS-WIDE dataset. The images from ImageNet are treated as the Image (**Im**) domain and the tags from NUS-WIDE as the Text (**Te**) domain. Also, each image and tag are characterized by $D_{4096}$ features and 64-D deep features, respectively. As our problem is under the MHDA setting, we introduce an Image-noise (**In**) domain, which is derived from the **Im** domain. Concretely, for each image in the **Im** domain, we first utilize PCA to reduce $D_{4096}$ features to 600-D PCA ($P_{600}$) features. Then, we add a 600-D Gaussian noise on the basis of $P_{600}$ features. Accordingly, **Im** and **In** are related, different, and heterogeneous. We construct one transfer task, i.e., **Im**, **In** $\rightarrow$ **Te**. For each source domain, we randomly pick up 100 images per class as the labeled samples. As for the target domain, we randomly select three tags per category as the labeled samples, and the remaining tags are used as the unlabeled samples.

*2) Implementation Details:* As we have a very limited number of labeled target samples, the cross-validation technique is not suitable for parameter selection. Thus, for a fair comparison, we empirically tune the hyperparameters of CWAN on the tasks of A ($D_{4096}$), D ($R_{2048}$) $\rightarrow$ W ($S_{800}$); Ar ($D_{4096}$), Cl ($R_{2048}$) $\rightarrow$ Re ($S_{800}$); and E, F $\rightarrow$ S, find a suitable hyper-parameter setting, and then apply it to other tasks. The details of network architecture and parameter setting of CWAN are given as follows. We implement the proposed CWAN based on the TensorFlow framework [74]. Each $g_{s_k}(\cdot)$ and $g_t(\cdot)$ are two-layer fully connected networks with the Leaky ReLU [75] activation function. $f(\cdot)$ is a one-layer fully connected network with the linear activation function. $d(\cdot)$ is a two-layer fully connected network with the ReLU [76] and linear activation functions, respectively. We optimize $\{f(\cdot), g(\cdot)\}$ and $d(\cdot)$ by utilizing the Adam optimizers [77] with learning rates of 0.004 and 0.001, respectively. We empirically tune the hyperparameters $\beta$, $\tau$, and $d_c$ from $\{0.01, 0.02, 0.03, 0.04, 0.05\}$, $\{0.001, 0.002, 0.003, 0.004, 0.005\}$, and $\{32, 64, 128, 256, 512, 1024\}$, respectively. We find that $\beta = 0.03$, $\tau = 0.004$, and $d_c = 256$ usually deliver the best performance for different tasks. Hence, such parameter settings are recommended as default settings for different real-world applications.

*3) Baselines:* We compare the proposed **CWAN** with thirteen state-of-the-art baseline methods. **SVMt** and **NNt** only take labeled target samples to train a support vector machine [78] and a neural network, respectively. **NNst** maps samples from all the domains into a common subspace by training a neural network with labeled samples. As noted in Section II, **DAMA** is the state-of-the-art MHDA method. Moreover, inspired by Xu *et al.* [25], Peng *et al.* [26], and Zhu *et al.* [27], we design two comparative scenarios for MHDA:

1) *Single-best:* It first decomposes the MHDA problem into multiple SHDA ones and then independently performs SHDA approaches on each problem to find the best result as the final result.
2) *Source-combine:* It first projects samples from all heterogeneous domains into a common subspace and then merges them into a larger source to perform SHDA approaches.

The *Single-best* is used to testify whether MHDA can outperform the best SHDA models, and the *Source-combine* is used for evaluating whether MHDA is necessary to exploit. To both ends, we adapt six state-of-the-art SHDA methods, i.e., CDLS [5], TNT [4], $DDA_L$ (DDA [33] with the *linear* projection functions), $DDA_N$ (DDA with the *nonlinear* projection functions), STN [6], and SSAN [38], into Single-best CDLS (**Sb-CDLS**), Single-best TNT (**Sb-TNT**), Single-best $DDA_L$ (**Sb-DDA$_L$**), Single-best $DDA_N$ (**Sb-DDA$_N$**), Single-best STN (**Sb-STN**), Single-best SSAN (**Sb-SSAN**), Source-combine $DDA_N$ (**Sc-DDA$_N$**), Source-combine STN (**Sc-STN**), and Source-combine SSAN (**Sc-SSAN**), for comparison. Since CDLS, TNT, and $DDA_L$ involve complex optimization solutions when new learnable parameters (i.e., feature transformations) are introduced, we do not adapt them into the source-combine scenario. Different from the above methods, $DDA_N$, STN, and SSAN are implemented with the deep learning framework (e.g., Tensorflow and PyTorch [79]), which supports automatic differentiation of the loss function [74]. Thus, $DDA_N$, STN, and SSAN can be easily adapted into the source-combine scenario. We choose the hyperparameters of the baseline methods in a similar manner to the CWAN.

In addition, as stated in Section II, *homogeneous* domain adaptation approaches cannot tackle heterogeneous inputs;

TABLE II
ACCURACY (%) ON THE *Multilingual Reuters Collection* DATASET FOR MHDA

| Method | E, F → S | E, G → S | E, I → S | F, G → S | F, I → S | G, I → S | Avg |
|---|---|---|---|---|---|---|---|
| SVMt [78] | | | 58.2±1.18 | | | | 58.2±1.18 |
| NNt | | | 59.78±1.18 | | | | 59.78±1.18 |
| NNst | 60.78±1.39 | 59.38±1.22 | 60.27±0.96 | 59.76±1.49 | 60.08±1.27 | 59.37±1.13 | 59.94±1.24 |
| DAMA [23] | 52.9±1.42 | 52.52±1.6 | 54.07±1.4 | 49.24±1.84 | 50.97±2.06 | 54.49±1.49 | 52.37±1.64 |
| Sb-CDLS [5] | 60.73±1.4 | 60.75±1.42 | 60.5±1.44 | 60.75±1.42 | 60.73±1.4 | 60.75±1.42 | 60.7±1.42 |
| Sb-DDA$_L$ [33] | 66.39±0.74 | 66.39±0.74 | 67.09±0.67 | 66.03±0.6 | 67.09±0.67 | 67.09±0.67 | 66.68±0.68 |
| Sb-DDA$_N$ [33] | 67.28±0.8 | 66.77±0.65 | 66.77±0.65 | 67.28±0.8 | 67.28±0.8 | 66.7±1.16 | 67.01±0.81 |
| Sb-STN [6] | 66.38±0.66 | 65.86±0.81 | 65.86±0.81 | 66.38±0.66 | 66.38±0.66 | 65.86±0.81 | 66.12±0.74 |
| Sb-SSAN [38] | 63.74±0.89 | 62.16±1.2 | 62.16±1.2 | 63.74±0.89 | 63.74±0.89 | 61.88±1.59 | 62.9±1.11 |
| Sc-DDA$_N$ [33] | 64.67±1.06 | 65.39±0.93 | 65.63±0.71 | 65.47±0.87 | 65.78±0.87 | 66.74±0.79 | 65.61±0.87 |
| Sc-STN [6] | 67.61±0.62 | 67.05±0.86 | 67.71±0.76 | 67.25±0.7 | 67.17±0.85 | 66.58±0.67 | 67.23±0.74 |
| Sc-SSAN [38] | 65.07±0.84 | 64.14±0.9 | 64.62±0.88 | 62.37±1.59 | 63.95±1.52 | 65.25±1.31 | 64.23±1.17 |
| CWAN | **70.4±1.01** | **70.17±0.82** | **71.45±0.64** | **70.49±1.02** | **71.57±0.88** | **70.79±0.93** | **70.81±0.88** |

thus, we do not include them for comparison. Also, the effectiveness of the transfer mechanisms in such approaches on heterogeneous benchmark datasets has not been verified. Although those transfer mechanisms perform well for the homogeneous problem, they may not be quite effective for *heterogeneous* problem since the latter is more complex and challenging. Investigating whether they are still effective under the heterogeneous scenarios is left for our future research.

*4) Evaluation Metric:* Following Tsai *et al.* [5] and Yan *et al.* [11], [29], we adopt the classification accuracy as the evaluation metric, which is calculated by

$$\text{Accuracy} = \frac{\sum_{\mathbf{x}_i^u \in \mathcal{D}_u} I(f(\mathbf{x}_i^u) = \mathbf{y}_i^u)}{n_u} \quad (12)$$

where $\mathbf{y}_i^u$ and $f(\mathbf{x}_i^u)$ are the ground-truth and predicted one-hot class labels of $\mathbf{x}_i^u$, respectively, and $I(\pi)$ is an indicator function taking the value of 1 if $\pi$ is true and 0 otherwise. For a fair comparison, we report the average classification accuracy with the standard error [80] of each method on ten random experiments. Also, in each random experiment, we record the classification accuracies of all the methods in the last iteration.

*B. Results*

*1) Results on the Multilingual Reuters Collection Dataset:* The results on the **Multilingual Reuters Collection** dataset are presented in Table II. We do not present the results of Sb-TNT here because its performance is much worse than the other approaches (e.g., 40.11% on the task of **E, F → S**). One possible conjecture is that the tree-based neural network may not be suitable to deal with reduced high-dimensional sparse features, resulting in overfitting. In addition, we also note that the original paper of TNT does not show the results on this dataset (please see details in [4]). From Table II, we can make several meaningful observations.

1) The proposed CWAN substantially outperforms all the baseline methods on all the transfer tasks. The average classification accuracy of CWAN is **70.81%**, which exceeds the shallow MHDA method, i.e., DAMA, and the best deep MHDA method, i.e., Sc-STN, by **18.44%** and **3.58%**, respectively. The results clearly validate the effectiveness and superiority of CWAN.

2) The performance of DAMA is very poor. This observation is similar to [12] because it utilizes only shallow structures and cannot differentiate the importance of multiple source domains.

3) All the methods with multiple sources except DAMA achieve comparable or better performance than the supervised learning methods, i.e., SVMt and NNt, which implies that these methods can produce effective transfer on those tasks.

4) CWAN, Sc-DDA$_N$, Sc-STN, and Sc-SSAN perform significantly better than NNst, which implies that the transfer mechanisms of those methods are both effective.

5) Sb-SSAN, Sb-STN, and Sb-DDA$_N$ perform better than Sb-CDLS. One important reason is that Sb-SSAN, Sb-STN, and Sb-DDA$_N$ are the deep approaches, while Sb-CDLS is a shallow one.

6) Sb-DDA$_N$ performs better than Sb-DDA$_L$, which implies that the nonlinear projection functions are more effective than the linear projection functions for seeking optimal common subspace.

7) The performance of Sb-CDLS, Sb-DDA$_L$, Sb-DDA$_N$, Sb-STN, and Sb-SSAN is worse than that of CWAN because they only utilize a single source domain for adaptation. The observation implies that only using a single best source domain is not a good strategy to address the MHDA problem.

8) CWAN performs better than Sc-DDA$_N$, Sc-STN, and Sc-SSAN. One important reason is that Sc-DDA$_N$, Sc-STN, and Sc-SSAN cannot characterize the importance of different source domains. In addition, the observation suggests that the MHDA is necessary to exploit.

*2) Results on the Office-Home Dataset:* The results on the **Office-Home** dataset are reported in Table III. We can summarize a number of insightful observations.

1) Again, the proposed CWAN performs the best. The average classification accuracy of CWAN is **52.16%**, which outperforms the shallow MHDA method, i.e., DAMA,

TABLE III

ACCURACY (%) ON THE *Office-Home* DATASET FOR MHDA

| Method | $D_{4096}, R_{2048} \rightarrow S_{800}$ | | | $S_{800}, R_{2048} \rightarrow D_{4096}$ | | | $S_{800}, D_{4096} \rightarrow R_{2048}$ | | | Avg |
|---|---|---|---|---|---|---|---|---|---|---|
| | Ar, Cl → Re | Ar, Pr → Re | Cl, Pr → Re | Ar, Cl → Re | Ar, Pr → Re | Cl, Pr → Re | Ar, Cl → Re | Ar, Pr → Re | Cl, Pr → Re | |
| SVMt [78] | | 10.72±0.22 | | | 46.44±0.47 | | | 82.16±0.3 | | 46.44±0.33 |
| NNt | | 11.44±0.25 | | | 46.89±0.4 | | | 80.42±0.36 | | 46.25±0.34 |
| NNst | 10.37±0.27 | 11.19±0.29 | 10.98±0.27 | 40.06±0.54 | 42.63±0.45 | 43.28±0.49 | 74.49±0.4 | 76.36±0.26 | 75.29±0.29 | 42.74±0.36 |
| DAMA [23] | 8.56±0.29 | 8.49±0.32 | 9.96±0.19 | 47.87±0.35 | 47.71±0.37 | 48.07±0.45 | 79.68±0.44 | 81.92±0.31 | 79.98±0.36 | 45.8±0.34 |
| Sb-CDLS [5] | 10.7±0.27 | 10.66±0.2 | 10.71±0.28 | 48.56±0.36 | 50.24±0.35 | 50.24±0.35 | 81.87±0.31 | 82.61±0.28 | 82.61±0.28 | 47.58±0.3 |
| Sb-TNT [4] | 11.29±0.23 | 11.29±0.23 | 10.82±0.3 | 51.71±0.28 | 53.28±0.49 | 53.28±0.49 | 82.62±0.21 | 83.08±0.34 | 83.08±0.34 | 48.94±0.32 |
| Sb-$DDA_L$ [33] | 11.12±0.24 | 11.12±0.26 | 11.12±0.26 | 47.35±0.83 | 48.71±0.68 | 48.72±0.68 | 80.08±0.32 | 80.08±0.32 | 79.33±0.46 | 46.4±0.45 |
| Sb-$DDA_N$ [33] | 11.32±0.32 | 11.61±0.3 | 11.61±0.3 | 49.86±0.59 | 51.29±0.34 | 51.29±0.34 | 82.83±0.32 | 84.11±0.2 | 84.11±0.2 | 48.67±0.32 |
| Sb-STN [6] | 10.8±0.25 | 10.89±0.29 | 11.01±0.29 | 51.87±0.41 | 53.09±0.58 | 53.09±0.58 | 83.35±0.33 | 83.91±0.26 | 83.91±0.26 | 49.1±0.36 |
| Sb-SSAN [38] | 12.27±0.24 | 12.27±0.24 | 12.19±0.25 | 49.93±0.36 | 51.58±0.34 | 51.58±0.34 | 86.08±0.29 | 86.02±0.14 | 86.02±0.14 | 49.77±0.26 |
| Sc-$DDA_N$ [33] | 11.24±0.3 | 11.19±0.24 | 11.23±0.25 | 49.12±0.77 | 49.4±0.79 | 51.05±0.39 | 81.18±0.28 | 82.65±0.4 | 81.77±0.31 | 47.65±0.41 |
| Sc-STN [6] | 10.5±0.33 | 11.29±0.28 | 10.86±0.33 | 43.17±0.3 | 46.56±0.35 | 43.97±0.26 | 76.16±0.34 | 79.31±0.32 | 74.79±0.44 | 44.07±0.33 |
| Sc-SSAN [38] | **13.09±0.26** | **12.45±0.3** | **12.28±0.29** | 45.55±0.56 | 49.79±0.46 | 46.65±0.34 | 85.37±0.28 | 85.56±0.19 | 84.43±0.27 | 48.35±0.33 |
| CWAN | 11.68±0.25 | 11.55±0.33 | 11.05±0.35 | **56.58±0.41** | **57.41±0.35** | **56.72±0.35** | **88.27±0.17** | **88.53±0.13** | **87.65±0.17** | **52.16±0.28** |

TABLE IV

ACCURACY (%) ON THE *Office-31* DATASET FOR MHDA

| Method | $D_{4096}, R_{2048} \rightarrow S_{800}$ | | | $S_{800}, R_{2048} \rightarrow D_{4096}$ | | | $S_{800}, D_{4096} \rightarrow R_{2048}$ | | | Avg |
|---|---|---|---|---|---|---|---|---|---|---|
| | A, D → W | A, W → D | W, D → A | A, D → W | A, W → D | W, D → A | A, D → W | A, W → D | W, D → A | |
| SVMt [78] | 55.63±0.75 | 54.05±0.85 | 23.86±0.35 | 80.33±0.54 | 80.22±0.87 | 56.4±0.75 | 93.4±0.36 | 95.01±0.35 | 84.81±0.31 | 69.3±0.57 |
| NNt | 55.33±0.84 | 54.69±1.25 | 24.74±0.41 | 79.13±0.8 | 79.16±0.68 | 55.94±0.71 | 93.62±0.37 | 95.7±0.36 | 84.96±0.33 | 69.25±0.64 |
| NNst | 51.4±0.68 | 50.27±0.84 | 22.43±0.32 | 79.86±0.33 | 80.86±0.56 | 52.24±0.5 | 92.14±0.46 | 94.94±0.42 | 83.1±0.38 | 67.47±0.5 |
| DAMA [23] | 55.03±1.14 | 45.83±1.78 | 11.96±0.6 | 79.19±0.92 | 78.62±0.91 | 53.91±0.89 | 91.23±0.94 | 94.81±0.47 | 83.15±0.69 | 65.97±0.93 |
| Sb-CDLS [5] | **60.64±1.04** | 57.68±0.76 | 26.04±0.47 | 82.58±0.81 | 81.63±0.78 | 57.22±0.56 | 94.99±0.42 | 96.49±0.29 | 85.24±0.41 | 71.39±0.62 |
| Sb-TNT [4] | 52.19±1.01 | 55.31±1.29 | 24.49±0.7 | **88.43±0.67** | **87.06±1.07** | 58.77±0.7 | 96.97±0.32 | **97.68±0.43** | 87.41±0.38 | 72.03±0.73 |
| Sb-$DDA_L$ [33] | 54.89±0.93 | 53.33±0.62 | 24.82±0.25 | 79.63±0.46 | 79.56±0.6 | 55.12±0.51 | 93.68±0.46 | 95.53±0.54 | 86.27±0.35 | 69.2±0.52 |
| Sb-$DDA_N$ [33] | 56.99±0.74 | 57.73±0.93 | 25.95±0.45 | 86.6±0.47 | 86.2±0.67 | 59.55±0.45 | 96.67±0.25 | **97.68±0.4** | 88.64±0.25 | 72.89±0.51 |
| Sb-STN [6] | 58.86±0.85 | 55.38±0.87 | 26.47±0.33 | 83.75±0.77 | 84.69±1.06 | 57.73±0.55 | 96.25±0.36 | 97.53±0.33 | 88.24±0.22 | 72.1±0.59 |
| Sb-SSAN [38] | 56.7±1.04 | **60.1±0.84** | **26.62±0.31** | 84.57±0.46 | 82.69±0.74 | 61.32±0.68 | 95.67±0.37 | 96.42±0.44 | 89.13±0.23 | 72.58±0.57 |
| Sc-$DDA_N$ [33] | 58.35±0.89 | 57.33±0.81 | 25.71±0.26 | 82.49±0.54 | 82.84±0.93 | 58.05±0.63 | 95.36±0.3 | 96.69±0.34 | 87.49±0.24 | 71.59±0.55 |
| Sc-STN [6] | 58.6±0.8 | 57.56±1.02 | 25.99±0.34 | 84.07±0.85 | 84.12±0.98 | 57.08±0.61 | 95.85±0.38 | 97.46±0.38 | 87.79±0.18 | 72.06±0.62 |
| Sc-SSAN [38] | 56.13±0.97 | 59.68±1.12 | 26.47±0.52 | 81.69±0.84 | 81.33±1.17 | 60.49±0.65 | 94.37±0.42 | 93.46±0.76 | 88.54±0.29 | 71.35±0.75 |
| CWAN | 59.67±0.95 | 58.77±0.91 | 25.77±0.38 | 87.07±0.68 | 85.53±0.66 | **62.75±0.53** | **97.59±0.2** | 97.65±0.47 | **90.03±0.17** | **73.87±0.55** |

and the best deep MHDA method, i.e., Sb-SSAN, by **6.36%** and **2.39%**, respectively. The results further corroborate the superiority of CWAN.

2) NNst performs worse than the supervised learning methods, i.e., SVMt and NNt. One possible reason is that the distributional divergence across the source and target domains may be large, resulting in negative transfer.

3) The *Source-combine* methods (i.e., Sc-$DDA_N$, Sc-STN, and Sc-SSAN) perform worse than the *Single-best* methods (i.e., Sb-$DDA_N$, Sb-STN, and Sb-SSAN). One reason is that there may be a large distributional divergence between source domains on those tasks, while the source-combine methods crudely combine all source domains into a larger source domain that hurts the performance.

4) Sc-STN is better than NNst but worse than SVMt and NNt. The observation implies that the transfer scheme of Sc-STN is effective, but it is not enough powerful to prevent negative transfer on those transfer tasks.

5) We have a similar observation as the *Multilingual Reuters Collection* dataset that CWAN is better than all the *Single-best* and *Source-combine* methods. The observation further verifies the necessity of exploiting the MHDA.

*3) Results on the Office-31 Dataset:* The results on the **Office-31** dataset are listed in Table IV. We have the following interesting observations.

1) The proposed CWAN yields the best performance in most settings. The average classification accuracy of CWAN is **73.87%**, which improves over the shallow MHDA method, i.e., DAMA, and the best deep MHDA method, i.e., Sb-$DDA_N$, by **7.9%** and **0.98%**, respectively. The results verify the superiority of CWAN again.

2) CWAN outperforms all the *Single-best* and *Source-combine* methods. The observations are similar to the results on the *Multilingual Reuters Collection* and *Office-Home* datasets.

*4) Results on the ImageNet + NUS-WIDE Dataset:* The results on the **ImageNet + NUS-WIDE** dataset are reported in Table V. We have the following interesting observations.

1) The proposed CWAN achieves the best performance on this dataset. The average classification accuracy of CWAN is **91.8%**, which improves over the shallow MHDA method, i.e., DAMA, and the best deep

TABLE V
ACCURACY (%) ON THE *ImageNet* + *NUS-WIDE* DATASET FOR MHDA

| Method | Im, In → Te |
|---|---|
| SVMt [78] | 84.9±0.51 |
| NNt | 84.95±0.4 |
| NNst | 81.56±0.73 |
| DAMA [23] | 77.77±1.11 |
| Sb-CDLS [5] | 87.38±0.47 |
| Sb-TNT [4] | 90.3±0.22 |
| Sb-DDA$_L$ [33] | 85.03±0.54 |
| Sb-DDA$_N$ [33] | 89.95±0.51 |
| Sb-STN [6] | 88.85±0.39 |
| Sb-SSAN [38] | 89.09±0.61 |
| Sc-DDA$_N$ [33] | 89.51±0.3 |
| Sc-STN [6] | 89.97±0.3 |
| Sc-SSAN [38] | 89.19±0.45 |
| CWAN | **91.8±0.32** |

MHDA method, i.e., Sb-TNT, by **14.03%** and **1.5%**, respectively. The results further corroborate the superiority of CWAN.

2) CWAN performs better than all the *Single-best* and *Source-combine* methods. We have the same observation as that on the *Multilingual Reuters Collection*, *Office-Home*, and *Office31* datasets.

## C. Analysis

*1) Ablation Study:* To delve deeper into the effectiveness of $\mathcal{L}_g$, $\ell_1$-norm used in $\mathcal{L}_g$, and $w_k$, we investigate several variants of CWAN: 1) CWAN (w/o $\mathcal{L}_g$), which removes $\mathcal{L}_g$ in (11); 2) CWAN ($\mathcal{L}_g = 0$), which leverages a shared second layer, i.e., $\{g_{s_k}^{l_2} = g_t^{l_2}\}_{k=1}^{K}$; 3) CWAN ($\mathcal{L}_g$ with $\ell_2$-norm), which adopts $\ell_2$-norm in (1); 4) CWAN ($w_k = 1$), which ignores the conditional weighting scheme by setting the weight of each source domain to be one; and 5) CWAN (w/o $\mathcal{L}_g \wedge w_k = 1$), which both ablates $\mathcal{L}_g$ and the conditional weighting scheme. Table VI shows the results on the *Office-31* dataset, which offers several insightful observations.

1) As expected, CWAN significantly outperforms its variants.
2) CWAN (w/o $\mathcal{L}_g$) and CWAN ($\mathcal{L}_g = 0$) are worse than CWAN, which indicates that $\mathcal{L}_g$ is helpful to further increase the performance; CWAN ($\mathcal{L}_g$ with $\ell_2$-norm) is worse than CWAN, which suggests that $\ell_1$-norm is more effective than $\ell_2$-norm for capturing the domain correlations.
3) CWAN ($w_k = 1$) is worse than CWAN, which implies that utilizing the conditional weighting scheme can further improve the performance; CWAN ($w_k = 1$) is worse than CWAN (w/o $\mathcal{L}_g$), which suggests that the conditional weighting scheme is more important than $\mathcal{L}_g$; CWAN (w/o $\mathcal{L}_g \wedge w_k = 1$) is worse than CWAN, CWAN (w/o $\mathcal{L}_g$), and CWAN ($w_k = 1$), which indicates that the conditional weighting scheme and $\mathcal{L}_g$ are both necessary and useful.
4) NNst is worse than CWAN (w/o $\mathcal{L}_g \wedge w_k = 1$), which implies the effectiveness of the adversarial learning strategy.

*2) Feature Visualization:* We adopt the t-SNE technique [81] to visualize the transformed samples on the task of **A** ($S_{800}$), **D** ($R_{2048}$) → **W** ($D_{4096}$). The visualization results are displayed in Fig. 5, which reveals several important observations.

1) From Fig. 5(a)–(c), we find that the discriminability of the three domains is ordered as follows: **D** ($R_{2048}$) > **W** ($D_{4096}$) > **A** ($S_{800}$). This is because the first two are represented by deep features, while the last one is represented by shallow features, and the deep features extracted by ResNet is better than those by DeCAF.
2) Fig. 5(d) shows that the transformed samples of different classes by DAMA are mixed together, which implies that DAMA does not match the class-conditional distributions well.
3) Fig. 5(e) shows that labeled samples are nicely grouped by NNst, while the unlabeled samples are distributed too closely to be differentiating.
4) Fig. 5(f)–(i) shows that Sc-DDA$_N$, Sc-STN, Sc-SSAN, and CWAN align the samples of source domains with those of the target domain nicely, and CWAN shows better discriminability.

*3) Weighting Evaluation:* We evaluate the effectiveness of the conditional weighting scheme on the task of **A** ($D_{4096}$), **D** ($R_{2048}$) → **W** ($S_{800}$). Fig. 6(a) shows the weights of different source domains w.r.t. the number of iterations. We can observe that their weights first decrease sharply, then increase rapidly, and, finally, decrease again until being stable. To better explain this, we depict the classification performance of all domains and the conditional distribution divergence between the source and target domains w.r.t. the number of iterations in Fig. 6(b) and (c), respectively. Note that, in Fig. 6(b), we present the classification performance on the labeled source samples and unlabeled target samples, respectively. Based on Fig. 6(b) and (c), we can explain the above observation as follows.

1) At the beginning of the optimization process, since the two optimization objectives in (11) can both minimize the weights of the source domains, the minimization on the weights is more frequent than that on the classification loss, which leads to the sharp decrease in weights and poor classification performance.
2) After the weights of the source domains drop to smaller values, the classification performance rapidly improves due to more minimization on the classification loss, which makes the samples change from disorder to order. In this process, the alignment of conditional distributions is difficult, which leads to the conditional distribution divergence increases, resulting in the rise in weights of the source domains.
3) After the weights of the source domains rise to larger values, the classification performance is gradually stable as the samples become more and more orderly. In this process, due to more minimization on the weights of the source domains, it leads to a decrease in their weights, resulting in the decline in the conditional distribution divergence.

In addition to the above observation, we also summarize the following insightful observations.

TABLE VI
ACCURACY (%) ON THE *Office-31* DATASET OF CWAN VARIANTS FOR MHDA

| Method | $D_{4096}, R_{2048} \to S_{800}$ | | | $S_{800}, R_{2048} \to D_{4096}$ | | | $S_{800}, D_{4096} \to R_{2048}$ | | | Avg |
|---|---|---|---|---|---|---|---|---|---|---|
| | A, D → W | A, W → D | W, D → A | A, D → W | A, W → D | W, D → A | A, D → W | A, W → D | W, D → A | |
| CWAN | **59.67±0.95** | **58.77±0.91** | 25.77±0.38 | **87.07±0.68** | **85.53±0.66** | **62.75±0.53** | **97.59±0.2** | **97.65±0.47** | **90.03±0.17** | **73.87±0.55** |
| CWAN (w/o $\mathcal{L}_g$) | 54.3±1.19 | 57.58±0.91 | 21.85±0.79 | 83.19±0.72 | 82.25±1.2 | 58.65±1.4 | 96.4±0.46 | 97.04±0.41 | 89.73±0.31 | 71.22±0.82 |
| CWAN ($\mathcal{L}_g = 0$) | 56.87±0.84 | 58.1±0.75 | **26.94±0.42** | 81.48±1.29 | 80.25±1.2 | 57.85±1.04 | 95.71±0.6 | 95.14±0.96 | 88.7±0.41 | 71.23±0.83 |
| CWAN ($\mathcal{L}_g$ with $\ell_2$-norm) | 58.09±0.98 | 56.37±0.97 | 24.97±0.47 | 80.41±0.93 | 79.8±0.84 | 60.88±0.92 | 94.13±0.87 | 94.77±0.63 | 88.2±0.94 | 70.85±0.84 |
| CWAN ($w_k = 1$) | 54.62±0.93 | 55.93±0.82 | 23.2±0.55 | 82.98±0.64 | 81.58±0.67 | 54.31±0.59 | 94.96±0.42 | 96.79±0.27 | 86.1±0.28 | 70.05±0.57 |
| CWAN (w/o $\mathcal{L}_g \wedge w_k = 1$) | 53.97±0.87 | 52.15±0.55 | 22.4±0.62 | 82.11±0.45 | 81.6±0.75 | 53.96±0.53 | 94.79±0.29 | 96.42±0.53 | 85.04±0.42 | 69.16±0.56 |
| NNst | 51.4±0.68 | 50.27±0.84 | 22.43±0.32 | 79.86±0.33 | 80.86±0.56 | 52.24±0.5 | 92.14±0.46 | 94.94±0.42 | 83.1±0.38 | 67.47±0.5 |

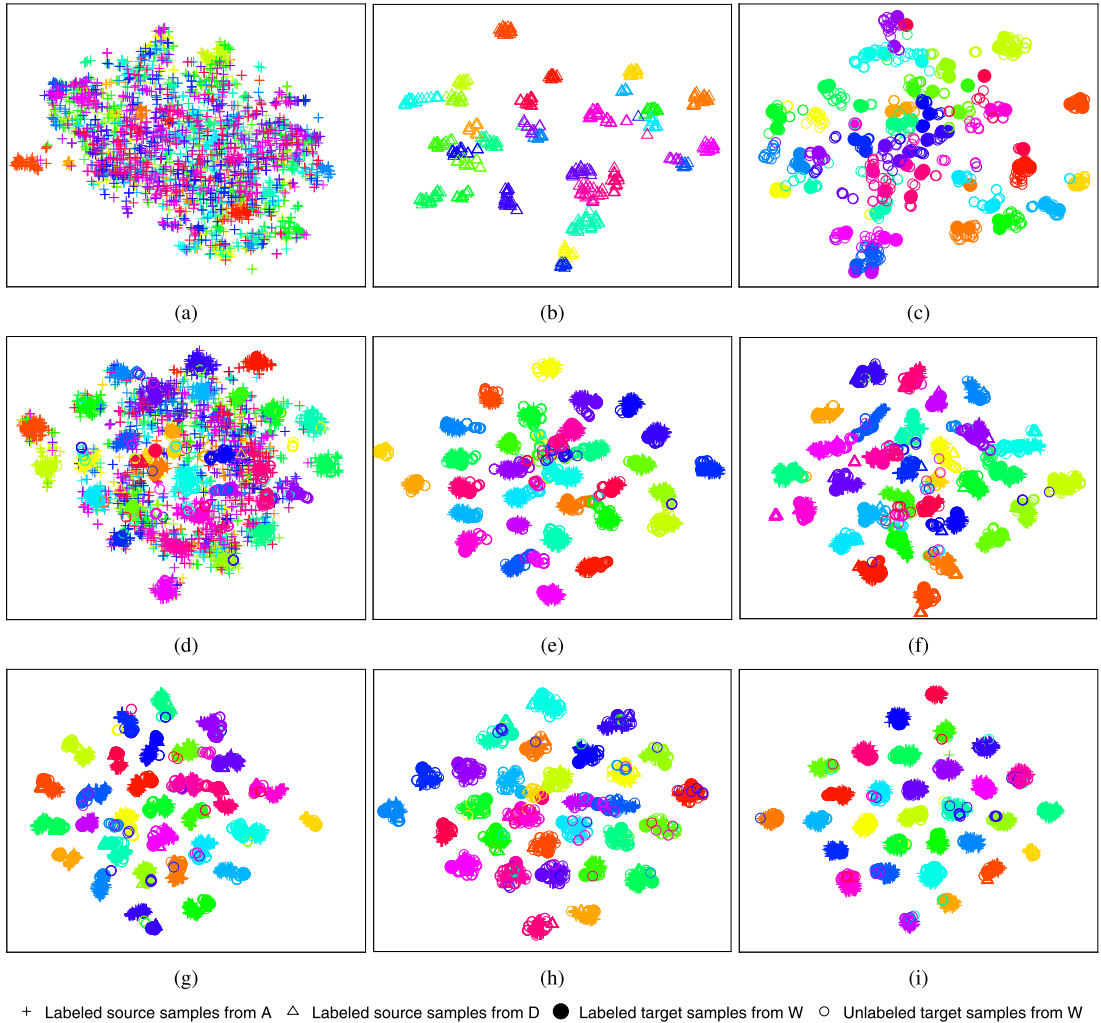

+ Labeled source samples from A    △ Labeled source samples from D    ● Labeled target samples from W    ○ Unlabeled target samples from W

Fig. 5. t-SNE visualization on the task of (a) A ($S_{800}$), (b) D ($R_{2048}$) → (c) W ($D_{4096}$). (a)–(c) Original feature representations of A ($S_{800}$), D ($R_{2048}$), and W ($D_{4096}$), respectively. (d)–(g) Learned feature representations of (d) DAMA, (e) NNst, (f) Sc-DDA$_N$, (g) Sc-STN, (h) Sc-SSAN, and (i) CWAN, respectively.

1) During the optimization process, the weights of **A** ($D_{4096}$) and **D** ($R_{2048}$) are distinct, which suggests that they have different contributions to **W** ($S_{800}$). Also, it also implies that the weighting scheme distinguishes the contributions of different source domains quite well.
2) At the end of the optimization process, the weights of **A** ($D_{4096}$) and **D** ($R_{2048}$) are small and similar, which indicates that they are both adapted to **W** ($S_{800}$) well.

3) The trend of their divergence curves is similar to that of their weight curves, which is reasonable since minimizing the weight will make the divergence small.
4) Comparing Fig. 6(a) with (c), we can observe that the weight of **A** ($D_{4096}$) is *greater* than that of **D** ($R_{2048}$), while the divergence between **A** ($D_{4096}$) and the target domain is *less* than that between **D** ($R_{2048}$) and the target domain. This observation indicates that a *smaller*

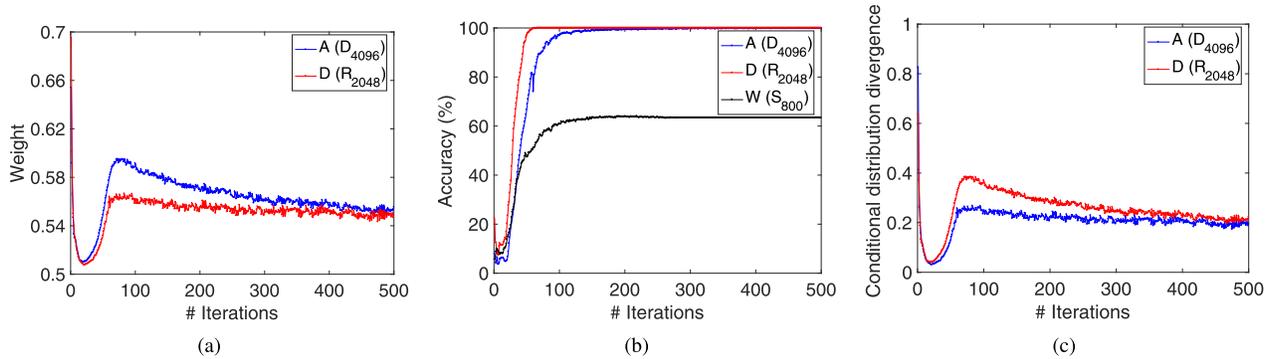

Fig. 6. Empirical analysis of the conditional weighting scheme on the task of **A** ($D_{4096}$), **D** ($R_{2048}$) → **W** ($S_{800}$). (a) Weight. (b) Accuracy. (c) Divergence.

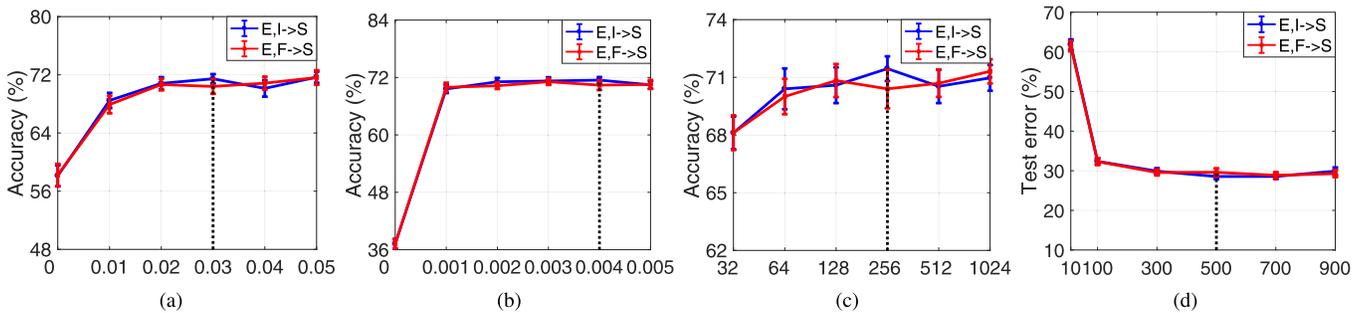

Fig. 7. Empirical analysis of parameter sensitivity and convergence on the tasks of E, I → S and E, F → S. (a) $\beta$. (b) $\tau$. (c) $d_c$. (d) # Iterations.

divergence leads to a *larger* weight, which obeys our weighting principle.

*4) Parameter Sensitivity and Convergence:* We analyze the parameter sensitivity and convergence on the tasks of **E, I → S** and **E, F → S**. Fig. 7(a)–(c) plots the accuracy *w.r.t.* distinct $\beta$, $\tau$, and $d_c$, respectively. We find that the default settings (i.e., $\beta = 0.03$, $\tau = 0.004$, and $d_c = 256$) lead to better performance on these tasks. Also, it is worth mentioning that CWAN achieves state-of-the-art performance on all the tasks with the default parameter settings. All the results indicate the stability and effectiveness of CWAN. In addition, as CWAN involves an alternative optimization procedure, we testify the convergence of CWAN with the test error. As shown in Fig. 7(d), the test errors first decrease gradually and then hardly change as more iterations are performed, which implies the convergence of CWAN.

*5) Noise Detection:* Since real-world applications may be very complex, we may collect some very dissimilar source domains. To further evaluate whether the conditional weighting scheme can detect very dissimilar source domains, we introduce a noise (**N**) domain generated by Gaussian distribution with random class labels. Note that the label space of **N** is consistent with that of **E**. Fig. 8 shows the weights of different source domains on the tasks of **E, F, N → S** and **G, I, N → S** in the last iteration. We can observe that the weights of **N** are the smallest on both tasks, which implies that the conditional weighting scheme detects very dissimilar source domains well.

*6) Multiple Sources:* To examine how the number of source domains $N_S$ affects the performance, we construct 11 three-class heterogeneous domains based on Gaussian distribution. The reason for constructing the synthetic dataset is that the *Office-31*, *Office-Home*, and *Multilingual Reuters Collection* datasets can at most construct a limited number of source domains (two, three, and four, respectively). Ten of them are used as source domains with dimensions ranging from 100 to 1000 by an increment of 100, and the rest 2000-D domain is viewed as the target domain. We change $N_S$ from zero to ten with a step size of two and present the results in Fig. 9. We can see that the accuracy first improves and then barely changes as $N_S$ increases. Also, the standard error first decreases monotonically and then tends to become stable with the increase in $N_S$. These observations suggest that the performance becomes more accurate and stable as more effective source domains are involved.

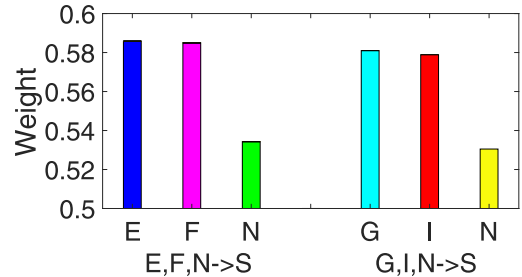

Fig. 8. Noise detection of the conditional weighting scheme.

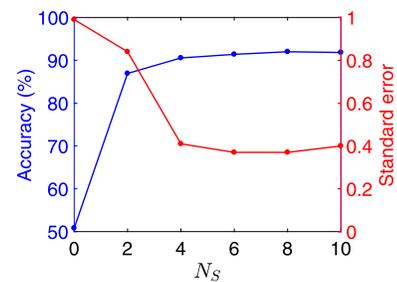

Fig. 9. Empirical analysis of the number of source domains.

## V. CONCLUSION

In this article, we propose a CWAN to address the MHDA problem, which alternatively learns a feature transformer, a label classifier, and a domain discriminator in an adversarial manner. A conditional weighting scheme is developed to not only weigh the source domains but also align the conditional distributions, which may inspire other researchers to design several powerful weighting schemes. Experiments on four real-world datasets verify the effectiveness of the CWAN. As a future direction, we intend to analyze the theoretical error bound for the CWAN under the MHDA setting. In addition, as noted in Section I, we believe that MHDA opens a new door for privacy protection. Thus, collecting and annotating large-scale multisource heterogeneous samples, especially for the privacy-protected samples, to form a benchmark database are also our future interests.

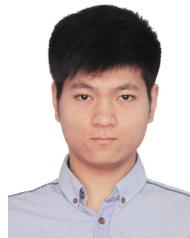

**Yuan Yao** is currently pursuing the Ph.D. degree with the Department of Computer Science and Technology, Harbin Institute of Technology, Shenzhen, China.

His current research interests include transfer learning, data mining, and machine learning.

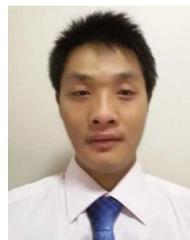

**Xutao Li** is currently an Associate Professor with the Department of Computer Science and Technology, Harbin Institute of Technology, Shenzhen, China. His research interests include data mining, machine learning, graph mining, and social network analysis, especially tensor-based learning and mining algorithms.

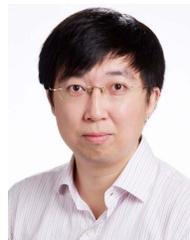

**Yu Zhang** (Member, IEEE) is currently an Associate Professor with the Department of Computer Science and Engineering, Southern University of Science and Technology, Shenzhen, China. His current research interests include machine learning and data mining, multitask learning, transfer learning, dimensionality reduction, metric learning, and semisupervised learning.

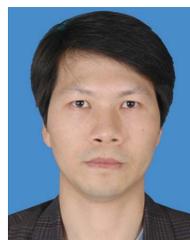

**Yunming Ye** is currently a Professor with the Department of Computer Science and Technology, Harbin Institute of Technology, Shenzhen, China. His research interests include data mining, text mining, and ensemble learning algorithms.